# Approximate Grammar for Information Extraction

**Akshar Bharati,**
**V.Sriram, B.Ravi Sekhar Reddy,**
**Rajeev Sangal**
**International Institute of Information Technology,**
**Hyderabad**
**{sriram,ravi_b}@gdit.iiit.net,{sangal}@iiit.net**

# Approximate Grammar for Information Extraction

## Abstract

In this paper, we present the concept of *Approximate grammar* and how it can be used to extract information from a document. As the structure of informational strings cannot be defined well in a document, we cannot use the conventional grammar rules to represent the information. Hence, the need arises to design an approximate grammar than can be used effectively to accomplish the task of *Information extraction*. Approximate grammars are a novel step in this direction. The rules of an approximate grammar can be given by a user or the machine can learn the rules from an annotated document. We have performed our experiments in both the above areas and the results have been impressive.

## Introduction

This paper proposes the idea of approximate grammars and its usage towards information extraction. An *informational string* is a sequence of words that convey definite information. For example: The string "India won by 4 wickets at Old Trafford" tells us about a match result. A document contains several informational strings that should be extracted. For example: a document on cricket would contain several informational strings like "the match would be held tomorrow at Lords", "the India-Pakistan match ended in a draw", "Sachin Tendulkar made 67 in 60 balls", etc.

The information extraction system consists of the following two stages,

1. Extraction of *atomic information* or *probe*. This atomic information consists of annotated texts, which fall into categories specific to the domain in consideration. For example: player-name, runs, wickets, venue, etc. could be the probes for Cricket.

2. Getting informational strings from the atomic information or *grouping*.

A document from which information is to be extracted is given to the probe, it first tags atomic information. (Sangal and Bansal, 2001)

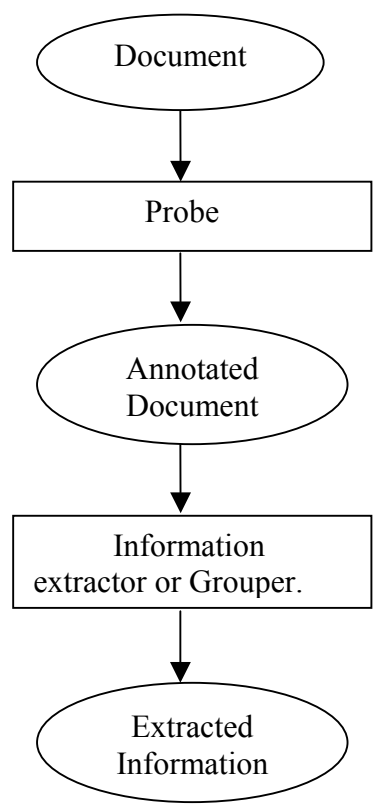

(Fig1. Information extraction system)

The second stage groups the atomic information into several groups, which represent the extracted information. To group these atomic units, we are required to know the structure of the document and have the knowledge of all the possible informational strings. The permitted informational strings can be captured by a grammar.

When a document is parsed according to the rules of the grammar, several parses are possible. A normal context-free grammar does not provide a mechanism to select the most appropriate parse from all the possible parses. To make that possible, we need to attach a probability with every rule of our grammar. Also, in an information string we can have redundant words, words that do not contribute to the informational string. These words can be called *noise*. In an informational string, apart from the normal atomic information or probes, we can also have patterns that can help us to extract the informational string. These patterns can be called atomic patterns.

In a document the informational units are usually close to each other. The maximum length of noise gives us the maximum possible distance between two informational units.

From all the above specifications that are useful to extract information, arises the need to have a new type of grammar. Approximate grammar incorporates all the above features and gives us the power, which can be used to do information extraction efficiently.

## Structure of Approximate Grammar

At this point, one could visualize approximate grammar as a set of probabilistic CFG rules with additional features, which are explained in the later part of the paper. The rule of an approximate grammar can take the following form.

```
non-terminal = terminals | non-terminals |
noise | probability of the rule

terminal: probe | string | root
```

"|" denotes OR.

## Features of Approximate Grammar

Probability of a Rule:

Probability associated with a particular rule helps the information extraction system to resolve the ambiguity that arises because of several possible parses.

For example, let's take a part of a document,

"<name> Sachin Tendulkar </name> made <runs> 55 runs </runs> in that match played at <location> Eden Gardens </location> .The match saw a total of <runs> 300 runs</runs> being made and <wickets> 5 wickets </wickets> falling. "

Say, we have two rules:

1> IMP -> {name} made {runs} {location}
2> IMP -> {location} saw {runs} {wickets}

IMP: Important informational string

The above rules would give us two parses.
The probabilities of the rules attached to each of the above rule would give us the score of the final parse. The parse with the highest score would be selected to give the appropriate informational strings.

Atomic strings or Words:

The atomic strings or words make a rule more specific and thus increases the precision with which an informational string is extracted.

For example, considering a part of the document:
"<name> Kapil Dev </name>  bowled <balls> 5 </balls>"

The above string would be extracted using both the following rules,

IMP -> {name} {balls}
IMP -> {name} bowled {balls}

But the second rule is stricter than the first rule and represents the information more precisely.

Noise:

Noise is defined as the pattern in an informational string, which has no relevance other than fact that it appears with the atomic units. So, between any two atomic units of an informational string, noise is permitted by the grammar.

For example:

Let, SN -> noise

Then, we give a rule as

IMP -> {name} SN bowled SN {balls}

We can specify the minimum and maximum lengths of the noise; this tells us how near or far can the two atomic units or atomic patterns in an informational string are. Roughly speaking, noise cushions the occurrence of the terminals and non-terminals of a grammar production.

So, if
 SN -> noise; minlength=0, maxlength=20

Then, there could be a maximum of 20 characters between {name} and "bowled" as well as between "bowled" and {balls} in the above rule as shown seperated by a semicolon.

Terminals:

The terminals can be divided into either of the following categories. The attribute 'cat' denotes the category.

a) Probe: Type of the annotated strings representing the atomic information.

   Example: <probe cat="player_name">

b) String: A particular string the user is expecting to be in the extracted information.

   Example: <string cat="cent" word="century">

c) Root: A special type of type "string" where the user can specify the root-word. Words in the target document having the morphological root as mentioned fall into this category.

   Example: <root cat="score" r-word="make" | "hit" | "score" >

Then the rule: *IMP-> player_name score cent*; where 'player_name', 'score' and 'cent' are the category names of the terminals as shown above, would capture the informational string,

"<player_name>Tendulkar</player_name> <score>made</score> a <cent>century</cent>".

Note: For the sake of simplicity, NOISE is not mentioned in the above rule. When NOISE is not mentioned, it is assumed to be there by the parser.

Scoring a Parse:

Each terminal is associated with a particular confidence with which it has been extracted.
Scoring a parse tree takes into account the probability of the rule and the scores of its child nodes.

Score of a terminal =
   Confidence with which a terminal is extracted.

Score of a node of a parse tree =
   Prob. Of rule * Average score of Child nodes

Example:

Given P-CFG's
   S-> NP VP ; prob=0.9
   NP -> det n ; prob=0.7
   VP -> v det n ; prob=0.6

And for the sentence

"The old man the boats"

Rules from the dictionary
   Det -> "the" ; conf= 0.9
   Adj -> "old" ; conf=0.6
   N -> "old" ; conf=0.3
   N -> "man" ; conf=0.7
   V -> "man" ; conf=0.25
   N -> "boats"; conf=0.8

Generated Tree

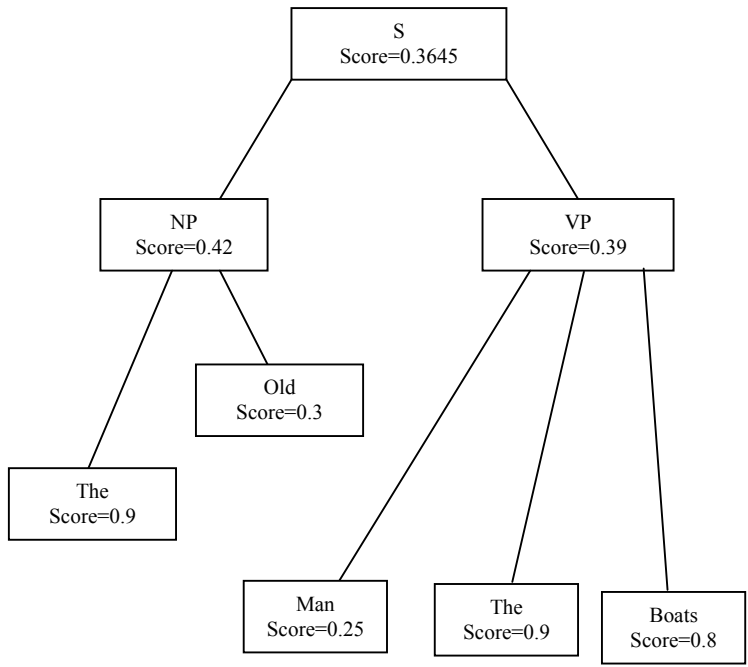

(Figure2. Generated Parse Tree for the given example)

**Advantages of using Approximate Grammars**

Resolving Ambiguity:

The score of a given parse gives the information extraction system, the power to select the appropriate parse from the several possible parses.

The following examples explain the above:

- Consider a piece of document,
  "B.A. Ambedkar is highly honored in India."

  The probes would tag the above in two possible ways.

  One interpretation is,
  "<degree> B.A</degree> Ambedkar is highly honored in <country> India </country>"

Note that the production,

IMP -> {degree} honored {country} ; is not a faulty production because it is valid for the following sentence: "<degree> M.Tech </degree> is honored in <country> India </country> .

Another interpretation is,

"<name>B.A.Ambedkar</name> is highly honored in <country>India</country>"

After training a sufficiently large corpus, one would find out that the probability of the production: IMP->{name}honored {country} would have a higher probability than the production: IMP->{degree}honored{country}

Then, the first interpretation of the probes is taken to be more appropriate, assuming that the confidence of extraction of atomic units is the same.

- Given the rules,

  IMP -> {country} {name}
        Prob= 0.1
  IMP -> {name} {country}
        Prob= 0.3

  Then for the piece of document,
  "<name> Sachin Tendulkar </name> plays for <country> India </country> that is being sponsored by <name> Mr. Sahara </name>".

  The informational string would be

  Sachin Tendulkar plays for India.

Proximity of Atomic Units:

In a document, the smaller units that build a bigger unit should be close to each other. The measure of maximum length of noise helps the system to find for units that are close to one another.

For example:

Given the rules,

SN -> noise; minlength=0, maxlength=10
IMP -> {name} SN {runs} SN {location}

And the piece of a document,

"<name> Sachin Tendulkar </name> made <runs> a duck </runs>. Hope he does well in <location> Calcutta </location>."

The above rule will not extract any string from the above piece of document because the length of noise between the runs and location is 22 while the maximum permitted distance is 10.

Machine Learning:

The machine can be used to learn the rules of approximate grammar by analyzing an annotated text. The probability of the rules and the maximum possible noise between any two atomic units is also learned. These learned rules are then applied to an unannotated text and the informational strings corresponding to the learnt rules are extracted.

## Implementation of Approximate Grammars

Chart Parsing: Chart Parsing is employed to parse the text according to the rules mentioned. A bottom-up approach is implemented. The algorithm of chart parsing provides an efficient way of obtaining multiple parses

The advantages of using Chart Parsing are:

- It avoids multiplication of effort.
- It provides a compact representation for 'local ambiguity'
- It provides a representation for 'partial parses'

Moreover, chart parsing provides a general framework in which alternative parsing and search strategies can be compared.

## Machine Learning to generate approximate grammars

In this part of the paper, we discuss how we achieved the generation of an approximate grammar by learning patterns from a set of tagged documents. Depending on the value of a parameter, the generated rules can be mapped to the extent of generality of the patterns exhibited in the document.

### a) Clustering of the patterns

As a primary step, the patterns of the informational strings given on the same sequence of their tags are clustered. Now, for each individual cluster, we aim to generate probabilistic grammar rules whose sum adds up to 1.

### b) Generating grammars for a specific cluster

Now that we have the set of informational strings following a specific pattern of occurrence of the atomic probes, we compute the TRI-data for the cluster. This data contains the TRI-frequency of every consecutive atomic probes. The TRI-frequency can be defined as the number of times the sub-pattern <probe1> "string" <probe2> has occurred in the cluster. The new set of stricter rules is generated by taking combinations of the various tag offsets found in the cluster. Note that , in the above mentioned general pattern, "string" can be null.

Consider the following example,
1. <IMP>   <name>Dravid</name> hit <runs>67 runs</runs> in the match   </IMP>.
2.   <IMP>   <name>Sachin</name>   made <runs>56</runs> before dawn </IMP>.
3.   <IMP>   <name>Laxman</name>   made <runs>34</runs> at the end of the innings </IMP>.

Corresponding TRI data.
<name> "hit" <runs>      - 1
<name> "made" <runs>     - 2
<runs> "match" </IMP>    - 1
<runs> "dawn" </IMP>     - 1
<runs> "innings" </IMP>  - 1

Generated CFG's

1. IMP -> <name> NOISE <runs> prob=0.5
2. IMP -> <name> "hit" <runs> "match" prob=0.0555
3. IMP -> <name> "hit" <runs> "dawn" prob=0.0555
4. IMP -> <name> "hit" <runs> "innings" prob=0.0555
5. IMP -> <name> "made" <runs> "match" prob=0.1111
6. IMP -> <name> "made" <runs> "dawn" prob=0.1111
7. IMP -> <name> "made" <runs> "innings" prob=0.1111

Note: For the sake of simplicity, NOISE is not mentioned in the above rule. When NOISE is not mentioned, it is assumed to be there by the parser.

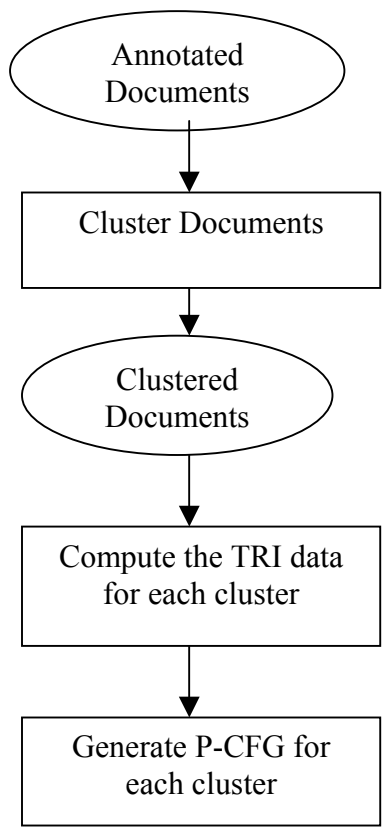

(Fig 3. Generating P-CFG's by machine learning)

RHO: An important parameter in P-CFG generation, which defines a threshold below which the observed patterns in question will be discarded because of their low frequency. Disposal of such patterns is important because these patterns give rise to ambiguous productions, thereby increasing the execution time.

We can extract the CFG's corresponding to the patterns observed based on a scale of generality. This can be done by filtering the TRI's above a threshold "RHO". Increasing the value of this parameter results in the generation of the patterns, which are more prevalent in the texts observed. Based on the tagged documents fed as input and the nature of the extracted information, we can fine-tune this parameter to achieve more concise and relevant results.

**Manually specifying the rules**

The user has the flexibility to express his idea of the information, which is relevant to him through this framework of approximate grammars. In this way, the user can instruct the machine to specifically extract a particular pattern with ease.

**Experiments and Results**

Experiments were carried out in the domain of Cricket. The texts were taken from **www.thatscricket.com** website. For obtaining the annotated texts, we have made use of filters which were developed on the cricket domain. The different filters used are : player_name, runs , wickets, balls , place, tournament, team. Training was done on 80%, and testing on the remaining 20%.

We were particularly interested in the outcome and other statistical details of the matches (individual scores of individual players, bowling performances of bowlers, partnerships, results etc.) and so we considered only the annotated parts where such information was present. Comments and other trivial informational strings regarding players were considered unimportant in our experimentation.

## Evaluation

Machine generated rules captured the essential information in the domain of cricket. Table 1. illustrates the rules specifying just the probes as non-terminals on the right-hand side of the production. Stricter versions of some of the rules might be having strings/roots as the terminals on the right hand side of the production and are listed at the end of the table 1. The probabilities of the rules mentioned are not stated because they assume different values depending on the value of RHO.

Note: For the sake of simplicity, NOISE is not mentioned between the probe non-terminals of the rules given below, but it is assumed to be there by the parser.

IMP: Extracted rules.

| |
|---|
| IMP-> team  team  runs  runs  tment  venue |
| IMP-> team player wickets runs player team runs |
| IMP-> player  runs  balls  balls  team  player |
| IMP-> runs  balls  player  wickets  player  player  wickets |
| IMP-> team  runs  wickets  player  runs  player  runs |
| IMP-> team  runs  team  venue |
| IMP-> team  venue  team  runs  player  wickets  runs  player  wickets  runs |
| IMP-> team  runs  player  runs  player  runs |
| IMP-> player  runs  runs  team  runs |
| IMP-> team  team  wickets  runs  player  runs  runs |
| IMP->  player  wickets  runs  player  wickets  runs  team  runs |
| IMP->  team runs wickets team  player runs |
| IMP-> player runs  player  runs  team  balls |
| IMP-> player runs  runs  player  runs  team  runs  wickets team  tment  venue |
| IMP-> player runs  team  wickets  player  player |
| IMP-> player  runs  player  runs  team  runs  runs  runs |
| IMP->  player team  tment  runs  balls  team  runs  team  runs |
| IMP->  player team  runs  wickets  player  player  wickets |
| IMP->  player  balls  runs |
| IMP->  player  runs  player  runs |
| IMP->  player  wickets |
| IMP-> player  "haul"  team  "beat"  team |
| IMP-> player  "help"  team  "beat"  team |
| IMP-> team  "regroup"  player  "bowl"  player  "ball" |
| IMP-> team  "regroup"  player  "paceman"  player  "ball" |
| IMP-> team  "open"  player  "bowl"  player  "ball" |
| IMP-> player  "hit"  runs  "keeper"  player  "wack"  runs |

(Table 1: Machine generated rules for the domain of cricket)

Evaluation Metric:

$$\text{Precision} = \frac{\text{Correctly extracted INF}}{\text{Total extracted INF}}$$

$$\text{Recall} = \frac{\text{Correctly extracted INF}}{\text{Total number of correct INF}}$$

INF: Informational Strings

| RHO | Precision | Recall |
|------|-----------|--------|
| 0.30 | 78 | 98.6 |
| 0.25 | 77.8 | 100 |
| 0.20 | 76.4 | 99.1 |
| 0.15 | 75.4 | 99.3 |

(Table 2: Results of the experiments)

Table.2 shows the results obtained after extracting the informational strings in the rest 20% of the corpus.

## Error Analysis

The misclassification of the probes in the training set brought down the precision to some extent.

Consider the piece of misclassified text:

<player_name> Hosts Hyderabad </player_name> bundled out <place> Kerala </place> at the <player_name>Gymkhana </player_name> grounds.

The above annotated text would generate the production:

IMP->player_name place player_name;

And eventually, the above production would extract the informational string: <player_name> Patel </player_name> stunning debut at the <place> Lords </place> came as a blow to <player_name> Nasser Hussain </player_name> .

The extracted informational string was not considered to be a valid one because we were specifically looking for outcome and other statistical details in a match played.

By increasing the value of RHO, we could eliminate the faulty productions generated after machine learning of the training set. Unfortunately, in this process we also lost some important productions because the frequency of some patterns in the training set which occurred in the testing set were very less. Cases where RHO value was less, some valid informational strings were later discarded in the parsing process because of their low score value in the parsing process. This is because another ambiguous competitor of that production had higher frequency in the training set.

## Conclusion

In this paper, we have proposed a framework to extract information efficiently and conveniently, considering the user's specifications and the loosely structured document in view. The machine learning algorithm to learn the rules of approximate grammars is giving good results as experimented.

## Future Work

We would like to categorize the informational strings into separate categories and train the system to obtain the Approximate Grammar rules for each of the categories. This would enable us to identify the category of the resultant Parse and thus would be more precise.

## Acknowledgement

Details to be given after blind reviewing is over.